\documentclass[runningheads]{llncs}
\usepackage{graphicx}
\usepackage{comment}
\usepackage{url}
\usepackage{subcaption}
\usepackage{calc}
\usepackage{rotating}
\usepackage{booktabs}
\usepackage{hyperref}
\usepackage{todonotes}
\usepackage{multirow, makecell}

\usepackage{tikz}
\usepackage[]{xcolor}

\newcommand{\marker}[3]{
  \tikz[baseline=(X.base)]{
    \node [fill=#1!40,rounded corners] (X) {#2:};
  }
  {\color{#1!80!black}#3}
}
\newcommand{\ma}[1]{\marker{blue}{MA}{#1}}  
\newcommand{\vds}[1]{\marker{green}{VdS}{#1}} 
\newcommand{\sa}[1]{\marker{brown}{SA}{#1}} 
\begin{document}
%



\title{Easy Semantification of Bioassays\thanks{Supported by TIB Leibniz Information Centre for Science and Technology, the EU H2020 ERC project ScienceGRaph (GA ID: 819536) and the ITN PERICO (GA ID: 812968).}}

%
%
\author{Marco Anteghini\inst{1,2}\orcidID{0000-0003-2794-3853} \and Jennifer D'Souza\inst{3}\orcidID{0000-0002-6616-9509}\and Vitor A.P. Martins dos Santos\inst{1,2}\orcidID{0000-0002-2352-9017} \and S\"oren Auer\inst{3}\orcidID{0000-0002-0698-2864}}
\authorrunning{Anteghini et al.}
%
\institute{Lifeglimmer GmbH, Markelstr. 38, 12163 Berlin, Germany
\and
Wageningen University \& Research, Laboratory of Systems \& Synthetic Biology, Stippeneng 4, 6708 WE, Wageningen, The Netherlands 
\email{\{anteghini,vds\}@lifeglimmer.com} \\
\and
TIB Leibniz Information Centre for Science and Technology, Hannover, Germany \\
\email{\{jennifer.dsouza,auer\}@tib.eu}}
\maketitle              
\begin{abstract} Biological data and knowledge bases increasingly rely on Semantic Web technologies and the use of knowledge graphs for data integration, retrieval and federated queries. We propose a solution for automatically \textit{semantifying biological assays}. Our solution contrasts the problem of automated semantification as labeling versus clustering where the two methods are on opposite ends of the method complexity spectrum. Characteristically modeling our problem, we find the clustering solution significantly outperforms a deep neural network state-of-the-art labeling approach. This novel contribution is based on two factors: 1) a learning objective closely modeled after the data outperforms an alternative approach with sophisticated semantic modeling; 2) automatically \textit{semantifying biological assays} achieves a high performance $F1$ of nearly 83\%, which to our knowledge is the first reported standardized evaluation of the task offering a strong benchmark model.

\end{abstract}
%
%
%

\section{Introduction}

Semantifying scholarly communication within the next-generation Knowledge-Graph-based Scholarly Digital Libraries, such as the Open Research Knowledge Graph\footnote{\url{https://www.orkg.org/}} (ORKG)~\cite{auer_soren_2018}, relies on core semantic techniques such as ontologized formalizations and Web resource identifiers~\cite{berners2001semantic}. This supports the mainstream \textit{Knowledge representation and reasoning} vision in AI. Further, semantified data can enable knowledge-based interoperability between multiple databases simply by reusing identifiers and utilizing no-SQL query languages such as SPARQL~\cite{sparql} that can perform distributed queries over the various data sources. Obtaining improved machine interpretability of scientific findings has seen keen interest in the Life Sciences~\cite{katayama2014biohackathon} domain. Many major bioinformatics databases such as UniProt~\cite{uniprot2021}, KEGG~\cite{kegg2000}, REACTOME~\cite{Jassal2019REACTOME} and the NCBI database~\cite{ncbi2017} which includes the PubChem BioAssay database now make their data available as Linked Data in which both biological entities and connections between them are ontologized with standardized relations and are identified through a unique identifier (an Internationalized Resource Identifier or IRI). In a parallel Computational Linguistics ecosphere, many recent interdisciplinary data collection and annotation efforts~\cite{chemrecipes,labprotocols,mysore2019materials,kuniyoshi2020annotating} are focused on the shallow semantic structuring of unstructured text based on the Life Sciences ontologies. E.g., instructional content in lab protocols, descriptions of chemical synthesis reactions, or bioassays. Thus information described otherwise in ad hoc ways within scholarly documents attain machine-actionable, structured representations. Such datasets inadvertently facilitate the development of automated machine readers.

In this work, we take up the problem of the automated semantification of Biological Assays (Bioassays). This problem has both Life Science-specific solutions as the Bioassay Ontology~\cite{visser2011BAO} and Computational Linguistics-based semantified unstructured text annotations~\cite{schurer2011bioassay,vempati2012formalizationBAO,clark2014fast}. A bioassay is, by definition, a standard biochemical test procedure used to determine the concentration or potency of a stimulus (physical, chemical, or biological) by its effect on living cells or tissues~\cite{hoskins1962uses,irwin1953statistical}. It is described with relevant information on basic procedures such as determining the signal that indicates biological activity, determining doses used during the test, calculation methods etc. Also, bioassays are always qualified and validated~\cite{little2019bioassaysection} to highlight their accuracy, repeatability, and adequacy for use in the measurement of relative potency. Thus, a semantic description of the assay represented as logical annotations consisting of property and value pairs is the semantic equivalent of the unstructured bioassay text. They would enable their large-scale analysis in diverse systems. Bioassay texts are semantified based on the BioAssay Ontology (BAO)~\cite{visser2011BAO,abeyruwan2014BAO}. The BAO describes chemical and biological screening assays and their related results to facilitate their categorization and data analysis. On the BioPortal\footnote{\url{https://bioportal.bioontology.org/}} where the BAO is hosted, the BAO showed 7513 classes and 227 properties dated June 3, 2021. Thus the semantification of an assay is a tedious human annotation task since they have to: 1) decide which ontologized class relation pair applies to a biossay; and 2) given a sentence from the bioassay text, decide whether it is expressible as a logical statement by the BAO. This results in a large decision space for the human annotator making it a time-consuming endeavor. Computational techniques fitted appropriately with the problem semantics can fully alleviate the tedious human annotation task.

\begin{figure}[!t]
  \centering
  \includegraphics[width=0.55\textwidth]{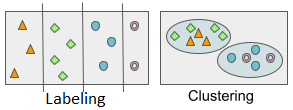}
  \caption{Illustration of labeling versus clustering to aggregate data points}
  \label{fig:labelvscluster}
\end{figure}

In this paper, we examine the computational aspects of the automated semantification of biological assays (bioassays) in light of two different approaches and their evaluations. We first formulate a labeling objective for bioassay semantification. This we recently proposed as a work-in-progress idea leveraging a transformer-based supervised classifier~\cite{anteghini2020representing,anteghini2020scibertbased}. Herein, we carry out in detail the experiments we began and further examine a novel clustering objective to bioassays semantification. Labeling and clustering are two methods of pattern identification used in machine learning. Although both techniques have certain similarities, the difference lies in the fact that labeling relies on a predefined set of labels assigned to objects, while clustering identifies similarities between objects, which it groups according to those characteristics in common and which differentiate them from other groups of objects. This is illustrated in \autoref{fig:labelvscluster}. On the one hand, we identify each logical statement of a semantified bioassay as a potential label. On the other hand, we observed that bioassays with similar text descriptions also had similar semantic representations. Thus a fine-grained clustering of the assays themselves could mean a cluster as a whole can be semantified by a standard set of labels. If it takes a classifier multiple passes to fully label an assay, it takes a clustering model just one pass over the data to semantify clusters. Via our experiments, we observed that labeling and clustering have contrasting score and time footprints. As a surprising result, the powerful transformer-based labeling method proves to be less accurate than a clustering solution at 54\% F1 vs. 83\% F1; and labeling with a large labels set has a significantly longer prediction time accounting for per-label classifications. 

In summary, the contributions of our work are:
\begin{enumerate}
    \item we formalize two machine learning objectives, i.e. labels classification and clustering, for the automated semantification of bioassays. Relatedly, we discuss the dataset characteristics and its adaptations. To our knowledge, these standardized machine learning tasks over a corpus of bioassays are discussed for the first time.
    \item we empirically evaluate the approaches and report unconventional findings that favor k-means clustering over the more resource-intensive transformers;
    \item we present an application of bioassay semantification within the Open Research Knowledge Graph scholarly contributions knowledge digitalization platform. The workflow allows scientists to upload bioassays, obtain automated semantified bioassays as results, and curate the semantic annotations. 
\end{enumerate}

\section{A Motivating Example for Bioassay Semantification}



\paragraph{\textbf{Assay ID 1960}} An example sentence from the assay is `Finally, fluorescence polarization can be used to effectively monitor the in vitro RNA-binding activity of both proteins using a standard fluorescence plate reader.' This sentence as it is is not computable. In other words, the terms `fluorescence polarization', `in vitro RNA-binding activity' or `standard fluorescence plate reader' in the unstructured text have no semantic interpretation to a computer. However, in the context of the standardized Bioassay terminology, the sentence is annotated with the following logical statement: `has detection method' $\rightarrow$ `fluorescence polarization' from the BioAssay ontology~\cite{visser2011BAO} grounded to the identifiers (bao:BAO\_0000207, bao:BAO\_0000003). This semantic annotation is now computable by machines, e.g., within reasoning tasks. But these annotations need to be manually curated by an expert who reads from context information in the phrase `Finally, fluorescence polarization can be used to effectively monitor' and is also familiar with the experimental setting of the assay. To semantify the above statement, the expert deduces that `high polarization' in `protein-probe complex' was detected by the method `fluorescence polarization.' However, making such decisions is an expensive human annotation task and nearly impossible at scale. Nevertheless, if such logical statements are annotated for a small set of bioassays, they can be easily annotated at scale via machine learning which is the focus of this work.

\section{Related Work}

\subsection{Corpora of Semantified Life Science Publications}

Increasingly, text mining initiatives are seeking out recipes or formulaic semantic patterns to automatically mine machine-actionable information from scholarly articles~\cite{chemrecipes,labprotocols,mysore2019materials,kuniyoshi2020annotating}. In~\cite{labprotocols}, they annotate wet lab protocols, covering a large spectrum of experimental biology, including neurology, epigenetics, metabolomics, cancer and stem cell biology, with actions corresponding to lab procedures and their attributes including materials, instruments and devices used to perform specific actions. Thereby the protocols then constituted a prespecified machine-readable format as opposed to the ad hoc documentation norm. Kulkarni et al.~\cite{labprotocols} release a large human-annotated corpus of semantified wet lab protocols to facilitate machine learning of such shallow semantic parsing over natural language instructions. Within scholarly articles, such instructions are typically published in the Materials and Method section in Biology and Chemistry fields. Along similar lines, inorganic materials synthesis reactions and procedures continue to reside as natural language descriptions in the text of journal articles. There is a growing need in such fields to find ways to systematically reduce the time and effort required to synthesize novel materials that presently remains one of the grand challenges in the field. In~\cite{chemrecipes,mysore2019materials}, to facilitate machine learning models for automatic extraction of materials syntheses from text, they present datasets of synthesis procedures annotated with semantic structure by domain experts in Materials Science. The types of information captured include synthesis operations (i.e. predicates), and the materials, conditions, apparatus and other entities participating in each synthesis step.

In this work, we leverage a similar semantically annotated corpus in the Life Science domain, but the knowledge theme tackled in our corpus is that of semantifying bioassays~\cite{visser2011BAO}. 
Normally, bioassays can be stored and accessed on PubChem~\cite{yanli2011pubchembio,wang2016pubchembioassay} which now contains more than 1.3M bioassays (22-06-2021). Only considering the period between 2015 and 2021, 389,835 new bioassays have been added to PubChem. To semantify a single bioassay is expert-specific and time-consuming. However, the process is not scalable for large-scale analyses, e.g. searching databases for related assays and comparisons or clustering similar entries. This requires the creation of new approaches to favor bioassays semantification, analysis, comparison and facilitate knowledge sharing. The ultimate goal would be to obtain a fully-automated software that can easily transform a human-friendly unstructured bioassay text report to a computer-friendly version as their semantic equivalent in the form of a set of logical statements.

\subsection{AI-based Scholarly Knowledge Graph Construction}

Early scholarly knowledge graph (SKG) construction initiatives were based on the sentences' information granularity. For this, ontologies and vocabularies were created~\cite{teufel1999annotation,soldatova2006,constantin2016,pertsas2017scholarly} from diverse aspects of the publication including discourse and specific themes as experiments; corpora were annotated~\cite{liakata2010corpora,fisas2016AMA}, and symbolic features-based ML techniques were implemented~\cite{liakata2012}. Recent scientific search technology led to new annotated corpora focusing on phrases with three or six types of generic scientific concepts in articles across up to ten different scholarly disciplines~\cite{dsouza2020stem,qzadeh2014acl,luan2018multi}, for which neural systems were developed~\cite{brack2020,ammar2017ai2,dessi2020}. In SKG creation, relation extraction has also raised keen interest, thanks also to community challenges such as ScienceIE 2017~\cite{scienceie}, SemEval 2018 Task 7~\cite{semeval-2018} and NlpContributionGraph 2021~\cite{ncg}, where participants tackled the problem of detecting semantic relations; newer advanced methods employed attention-based bidirectional long short-term memory networks (BiLSTM)~\cite{zhou2016attention} or used dynamic span graph framework based on BiLSTMs~\cite{wadden2019EntityRA}. Recently, strategically designed neural-symbolic hybrid approaches have proven effective~\cite{uiuc}.

For the scholarly knowledge theme of structuring Bioassays, specifically, the only prior machine learning approach was a morpho-syntactic features-based Bayes classifier~\cite{clark2014fast}. This early system, however, had unreplicable human-engineered aspects and non-standard evaluations. We focus on machine learning that entails no additional hand-engineering and report standardized evaluations.

\begin{table}[!tb]
\small
\centering
\begin{tabular}{l} \\\toprule
\textsc{has participant} $\rightarrow$ \textsc{DMSO} \\
\textsc{has assay phase characteristic} $\rightarrow$ \textsc{homogeneous phase}\\
\textsc{has temperature value} $\rightarrow$ \textsc{25 degree celsius} \\
\textsc{has incubation time value} $\rightarrow$ \textsc{20 minute} \\
\bottomrule
\end{tabular}
\caption{\small{Four example logical statements (from 50 total) for the semantified PubChem Assay with ID 360 (\url{https://pubchem.ncbi.nlm.nih.gov/bioassay/360}). Note, these statements are triples with subject `Bioassay.'}}
\label{tab:statementseg}
\end{table}

\section{Materials and Methods}

\subsection{An Expert-Annotated Semantified Bioassays Corpus}

To develop our automated semantifiers, we leverage a corpus comprising an expert-annotated collection of 983 semantified bioassays~\cite{schurer2011bioassay,vempati2012formalizationBAO}. In \autoref{tab:statementseg}, we show four logical statements of a semantified bioassay (ID 360 in PubChem) as an example. Each logical statement is expressed as a predicate and value pair. In the chosen example, the first two statements are \textit{ontologized} statements, i.e. the predicate and value pair are in the Bioassay Ontology (BAO)~\cite{visser2011BAO}. These annotations are made by a domain expert based on comprehensive knowledge of the BAO which contains thousands of predicate value pairs as semantification candidates. The next two statements are \textit{partially ontologized}, i.e. their predicates can be found in the BAO but the values are directly from the bioassay text description and hence are bioassay-specific. These statements report the various specific measurements made in the course of the bioassay. Semantified Bioassays contain both \textit{ontologized} and \textit{partially ontologized} statements. For the semantification task addressed in this paper, we restrict ourselves only to the \textit{ontologized} statements of each semantified bioassay. For this, we prune all \textit{partially ontologized} statements from each semantified assay. In \autoref{tab:dataset_statistics}, we summarize the dataset statistics for the original corpus with all the statements and the corpus we use after pruning. We can see that prior to pruning, the original corpus had 5524 total unique statements overall, which after pruning are reduced to 1906 statements. In the pruned corpus, bioassays have between 2 minimum and 87 maximum statements at an average of 37 statements. Considering only the predicates in these 1906 total statements, some predicates apply to semantify a bioassay more commonly than others. This is shown via the predicates statistics reported in \autoref{tab:methods:dataset}. In particular, 94\% of the semantic statements comprise only the 40 most commonly occurring predicates from a total of 80 unique predicates. Note this labels repetition detail of the corpus is critical since the labeling of bioassays with semantic statements are only among those observed in the data.

\subsubsection{Corpus Formalization.} 
Let $B$ be the overall semantified bioassays collection. A bioassay $b$ from $B$ is semantified with a set of \textit{ontologized} logical statements $sls$ (or semantic statements) which is $sls = \{ls_1, ls_2, ls_3, ..., ls_k\}$ where $ls_x$ is a logical statement $\in LS$ such that $LS$ is the collection of all the distinct \textit{ontologized} logical statements used for semantification seen in the training data. And $sls$ has $k$ different statements when taken together form the semantic equivalent of bioassay $b$. Across bioassays, their corresponding $sls$ sizes vary. 
As shown in \autoref{tab:dataset_statistics}, the corpus we use has $|LS| = $ 1906 unique statements  (after pruning the \textit{partially ontologized} statements).

\begin{table}[!tb]
    \centering
    \begin{tabular}{p{2cm}|p{2cm}|p{2cm}|p{2cm}|p{2cm}}\toprule
              & AVERAGE & MINIMUM & MAXIMUM & TOTAL \\\midrule
    original  & 56 & 7 & 162 & 5524 \\
    pruned    & 37 & 2 & 87 & 1906 \\\bottomrule
    \end{tabular}
    \caption{Semantified bioassays corpus statistics shown before (`original' row) and after (`pruned' row) pruning its \textit{partially ontologized} statements. \small{Note the corpus used for the work in this paper is the `pruned' version.}}
    \label{tab:dataset_statistics}
\end{table}

\begin{table}[!tb]
    \centering
    \begin{tabular}{p{1.35cm}|p{1.35cm}|p{1.35cm}|p{1.35cm}|p{1.35cm}|p{1.35cm}|p{1.35cm}|p{1.35cm}}
    \multicolumn{4}{l}{} & \multicolumn{4}{l}{} \\\toprule
         top 10 & top 20 & top 30 & top 40 & top 50 & top 60 & top 70 & top 80 \\\midrule
         795 & 959 & 1492 & 1804 & 1866 & 1879 & 1896 & 1906 \\
         (41.7) & (50.3) & (78.3) & (94.6) & (97.9) & (98.6) & (99.5) & (100.0) \\
         \bottomrule
    \end{tabular}
    \caption{Fine-grained pruned semantified corpus statistics in terms of the top 10, 20, 30, etc., most common predicates seen in the statements. E.g., the top 10 column contains the ten most frequently occurring predicates in the 1906 statements. Note the last column (`top 80') reflects the total unique predicates in the corpus. The rows show the number of the unique statements with the corresponding frequent predicates. The parenthesized numbers show the statements' proportion in the overall corpus.}
    \label{tab:methods:dataset}
\end{table}

Two semantification machine learning objectives are contrasted next.

\subsection{\textit{Labeling} Task Definition for Bioassay Semantification}
Bioassays semantification can be addressed as a labeling problem. In this scenario, each logical statement can be treated within a binary classification task as applicable or not. On average in our data, a bioassay could then have around 37 applicable logical statements from $LS$. The task can be formalized as follows.

\subsubsection{Task Formalism.} Each input data instance is the pair $(b,ls; c)$ where $c \in \{true, false\}$ is the classification of the label $ls$. Thus, specifically, our semantification problem entails classifying labels: $(b,ls)$ is $true$ if $ls \in$ logical statements set of $b$, else $false$. The $false$ instances are formed by pairing $b$ with any other label not in the logical statements set $sls$ of $b$.

Intuitively, this task formulation is meaningful because it emulates the way the human expert annotates the data. Basically, the expert, from their memory of all logical statements $LS$, simply assigns $ls$ to a given $b$ if they deem it as $true$; irrelevant statements are not considered, thus implicitly deemed $false$.

\subsubsection{Task Model.} Our machine learning system is the state-of-the-art, bidirectional transformer-based SciBERT~\cite{beltagy2019scibert}, pre-trained on millions of scientific articles. 
For bioassay semantification, we use the SciBERT classification architecture. In each data instance $(b,ls; c)$, the classifier input representation for the pair `$b,ls$' is the standard SciBERT format, treating them as sentence pairs separated by the special [SEP] token; the special classification token ([CLS]) remains the first token of every instance. Its final hidden state is used as the aggregate sequence representation for classification and is fed into a linear classification layer.

\subsection{\textit{Clustering} Task Definition for Bioassay Semantification}

We define clustering as the second machine learning strategy. This is from corpus observations wherein bioassays with similar text descriptions were semantified with similar sets of logical statements. Thus, bioassays could be clustered based on their text descriptions into semantic groups and each cluster group could be collectively semantified for its bioassays. This task formalism is as follows.

\subsubsection{Task Formalism.} Let $K$ be the total number of clusters of bioassays represented by the set $C = \{c_1, c_2, ..., c_K\}$. $B_{train} = \{b_1, b_2, ..., b_n\}$ corresponds to the total bioassays in the training set used to obtain optimal cluster centroids; and $V_{train} = \{v_1, v_2, ..., v_n\}$ is the vectorized representation of each bioassay to fit the clustering model. Note, $K < n$. Further, each cluster $c_x$ is associated with all the distinct logical statements of the bioassays in the respective cluster group. If cluster $c_x$ is fitted with two bioassays $b_p$ and $b_q$ in the training set, then $c_x$ is associated with $sls_{c_{x}} = sls_{b_{p}} \cap sls_{b_{q}}$. Thus, new logical statements sets are formed as $\{ls_{c_1}, ls_{c_2}, ..., ls_{c_K}\}$ associated with the $K$ clusters. After the clustering semantification model is fitted with $V_{train}$, semantification is performed. Each new bioassay $b_{test}$ is assigned based on $v_{test}$ to its closest cluster and semantified with the logical statements set of that cluster.

Clustering has the following alternative semantification task intuition. The domain expert tries to repeat their semantification decisions as much as possible based on similar bioassays they already annotated. In other words, for a new bioassay, they would copy as many logical statements from a similar already semantified bioassay and then decide if additional logical statements were needed. While this latter aspect is not modeled within the clustering problem, our results show that just copying the logical statements between similar bioassays is a significantly accurate automatic semantification strategy.

\subsubsection{Task Model.} Each bioassay text is represented based on the TF-IDF~\cite{tfidf} vectorized format. The clustering approach we employ is the K-means algorithm~\cite{kmeans}. To determine the optimal clusters $K$, we employ the elbow optimization strategy that tries to select the smallest number of clusters accounting for the largest amount of variation in the data~\cite{syakur2018integration}.\footnote{\url{https://scikit-learn.org/stable/modules/generated/sklearn.cluster.KMeans.html}}

\section{Bioassay Semantification Experiments}

\subsection{Experimental Setup}

\textbf{1. Labeling Task-specific Settings.} Unlike clustering, the labeling task entails defining \textit{false} logical statement semantification candidates as well. Since each assay had on average 37 \textit{true} logical statements, we experimented with a random set of \textit{false} (RF) statements in the range between 100 and 200 in increments of 10. The values were set to avoid biasing the classifier on only \textit{false} inferences but also to be sufficiently representative. \textbf{2. Three-fold Cross Validation.} For both labeling and clustering, we performed 3-fold cross validation experiments with a training/test set distribution of 600 and 300 assays, respectively. The test set assays were selected such that they were unique between the folds. \textbf{3. Evaluation Metrics.} We measure the standard precision, recall, and F1 scores for bioassay semantification per fold experiment. The final scores are then averaged over the three folds.


\begin{table}[!tb]
    \centering
    \begin{tabular}{p{2cm}|p{2cm}|p{2cm}|p{2cm}}
    \toprule
         & P & R & F1
         \\\midrule
         160RF & 0.33 & 0.94 & 0.49\\
         \textbf{170RF} & \underline{\textbf{0.37}} & \underline{\textbf{0.94}} & \underline{\textbf{0.54}} \\
         180RF & 0.35 & 0.94 & 0.51 \\\bottomrule
    \end{tabular}
    \caption{Bioassay semantification results by SciBERT-based labels classification. The first column shows the number of $false$ statements (RF) that each bioassay was labeled with---the rows report 3 different experiments (170RF as optimal).}
    \label{tab:results:SCIBERT:fulldataset}
\end{table}

\begin{table}[!tb]
    \centering
    \begin{tabular}{c|c|c|c|c|c|c|c|c|}
      &top10 & top20 & top30 & top40 & top50 & top60 & top70 & full \\\midrule
      TPU &   34s & 38s & 42s & 1m 10s & 1m 24s & 1m 22s & 1m 12s & 1m 20s\\
      CPU & 28m 10s & 29m 15s & 34m 7s & 58m 3s & 1h 6m & 1h 6m 14s & 1h 6m 4s &1h 6m 8s \\\bottomrule 
    \end{tabular}
    \caption{Rate of semantifying bioassays on various corpus subsets using SciBERT}
    \label{tab:time}
\end{table}

\begin{table}[!tb]
\centering
\begin{tabular}{p{1.7cm}|p{0.9cm}p{0.9cm}p{0.9cm}|p{0.02cm}|p{1.7cm}|p{0.9cm}p{0.9cm}p{0.9cm}} \toprule
   predicates    & P & R & F1 &  &  predicates  & P & R & F1 \\ \midrule
\textit{top 10} & \underline{0.53} & 0.94 & \underline{0.67} &  & \textit{top 50} & 0.36 & 0.95 & 0.52  \\
\textit{top 20} & 0.50  & 0.89 &  0.64  &  & \textit{top 60} & 0.41 & 0.92 & 0.57 \\
\textit{top 30} & 0.45  & \underline{0.95}  & 0.61 &  & \textit{top 70} & 0.32 & 0.95 & 0.48 \\
\textit{top 40} & 0.37  &  0.94 &  0.52  &  & \textit{all 80} & \textbf{0.37} & \textbf{0.94} & \textbf{0.54}  \\\bottomrule
\end{tabular}
\caption{SciBERT-based bioassay semantification on corpus subsets starting with only the statements containing the 10 most common predicates (\textit{top 10} row) until the full corpus (\textit{all 80} row). \small{In these experiments, the optimal 170RF was used.}}
\label{tab:results:SCIBERT:top}
\end{table}

\subsection{Experimental Results}

\subsubsection{SciBERT-based Semantification.} Given the results in \autoref{tab:results:SCIBERT:fulldataset}, we examine the \textit{\textbf{RQ}: is the proposed transformer-based neural method effective at semantifying bioassays?} A score of 0.54 $F1$ tells us, suprisingly, that our attempted neural-based method is not an effective solution to the problem which is a surprising result since it is the state-of-the-art in classification tasks over scientific data~\cite{beltagy2019scibert}. Further, it proves practically inefficient, since, given the full corpus of statements, each test assay is semantified at a rate of 1 hour on the CPU (see \autoref{tab:time}). On smaller subsets of the statement labels, the time is indeed faster and the scores are better (see \autoref{tab:results:SCIBERT:top}), however, time performance rate of 28 minutes on the smallest subset is still impractical.

\begin{table}[!tb]
\begin{tabular}{p{1.2cm}|p{0.65cm}p{0.65cm}p{0.95cm}|p{0.65cm}p{0.65cm}p{0.95cm}|p{0.65cm}p{0.65cm}p{0.95cm}|p{0.65cm}p{0.65cm}p{0.95cm}|p{0.65cm}p{0.65cm}p{0.95cm}}\toprule
\multirowcell{2}{Num. of \\ Clusters} & \multicolumn{3}{l|}{Labels freq $\ge$ 5} & \multicolumn{3}{l|}{Labels freq $\ge$ 4} & \multicolumn{3}{l|}{Labels freq $\ge$ 3} & \multicolumn{3}{l|}{Labels freq $\ge$ 2} & \multicolumn{3}{l}{Labels freq $\ge$ 1} \\
& P & R & F1 & P & R & F1 & P & R & F1 & P & R & F1 & P & R & F1   \\ \hline
50  & 0.54 & \underline{\textbf{0.75}} & \underline{\textbf{0.63}} & 0.48 & \underline{\textbf{0.80}} & 0.60 & 0.40 & \underline{\textbf{0.84}} & 0.54 & 0.32 & \underline{\textbf{0.89}} & 0.47 & 0.19 & \underline{\textbf{0.94}} & 0.31      \\
100 & 0.69 & 0.59 & 0.63 & 0.66 & 0.66 & \underline{\textbf{0.66}} $\uparrow$ & 0.62       & 0.76       & 0.68 $\uparrow$      & 0.53       & 0.85       & 0.66  $\uparrow$    & 0.32       & 0.92       & 0.47  $\uparrow$     \\
150 & 0.83 & 0.40 & 0.54 $\downarrow$ & 0.80 & 0.49 & 0.61 $\downarrow$ & 0.76 & 0.63       & \underline{\textbf{0.69}} $\uparrow$  & 0.70 & 0.79 & \underline{\textbf{0.74}} $\uparrow$ & 0.54 & 0.90 & 0.68   $\uparrow$    \\
200 & 0.86       & 0.34       & 0.49 $\downarrow$ & 0.83       & 0.43       & 0.56  $\downarrow$   & 0.80       & 0.56       & 0.66 $\downarrow$     & 0.76       & 0.72       & 0.74      & 0.66       & 0.89       & 0.75  $\uparrow$     \\
250 & 0.88       & 0.22       & 0.36 $\downarrow$ & 0.86       & 0.31       & 0.45  $\downarrow$    & 0.85       & 0.44       & 0.58 $\downarrow$     & 0.79       & 0.65       & 0.72  $\downarrow$  & 0.71       & 0.88       & 0.79  $\uparrow$     \\
300 & 0.91       & 0.18       & 0.30 $\downarrow$ & 0.88       & 0.24       & 0.37   $\downarrow$   & 0.86       & 0.35       & 0.50  $\downarrow$    & 0.81       & 0.56       & 0.66   $\downarrow$   & 0.75       & 0.86       & 0.80    $\uparrow$  \\
350 & 0.94       & 0.10       & 0.17 $\downarrow$ & 0.90       & 0.15       & 0.25  $\downarrow$    & 0.88       & 0.27       & 0.41  $\downarrow$    & 0.84       & 0.47       & 0.60  $\downarrow$    & 0.78       & 0.86       & 0.82    $\uparrow$   \\
400 & 0.93       & 0.06       & 0.11 $\downarrow$ & 0.93       & 0.09       & 0.17  $\downarrow$    & 0.91       & 0.20       & 0.32  $\downarrow$    & 0.86       & 0.38       & 0.53  $\downarrow$    & 0.80       & 0.85       & 0.82      \\
450 & 0.95       & 0.05       & 0.10 $\downarrow$ & 0.94       & 0.08       & 0.14  $\downarrow$    & 0.93       & 0.12       & 0.22  $\downarrow$    & 0.86       & 0.27       & 0.41   $\downarrow$   & 0.81       & 0.85       & \underline{\textbf{0.83}}     $\uparrow$  \\
500 & 0.95       & 0.03       & 0.06 $\downarrow$ & 0.94       & 0.05       & 0.09 $\downarrow$     & 0.93       & 0.08       & 0.15  $\downarrow$    & 0.88       & 0.17       & 0.28  $\downarrow$    & 0.82       & 0.85       & 0.83      \\
550 & 0.95       & 0.03       & 0.06 $\downarrow$ & 0.95       & 0.03       & 0.06 $\downarrow$     & 0.94       & 0.04       & 0.08   $\downarrow$   & 0.89       & 0.09       & 0.17  $\downarrow$    & 0.82       & 0.84       & 0.83      \\
600 & \underline{\textbf{1.0}} & 0.02 & 0.05 $\downarrow$ & \underline{\textbf{0.95}} & 0.02 & 0.05 $\downarrow$ & \underline{\textbf{0.96}} & 0.03 & 0.06 $\downarrow$ & \underline{\textbf{0.94}} & 0.04 & 0.07 $\downarrow$ & \underline{\textbf{0.83}}       & 0.84       & 0.83\\\bottomrule     
\end{tabular}
\caption{Bioassay semantification results by K-means clustering}
\label{clustering-results}
\end{table}

\subsubsection{K-means Clustering-based Semantification.} Detailed results with their performance rise and fall trends are shown in \autoref{clustering-results} for different cluster sizes and labels frequency thresholds within the clusters. E.g., the `Labels freq $\geq$ 5' column evaluates only the statements that appeared 5 or more times within the cluster groups when the semantic statements from the various bioassays were aggregated. As the labels frequency threshold is lowered, the semantification score rises. The best scores are obtained when all the statements are considered (the `Labels freq $\geq$ 1' column). This method obtains a high semantification score of 0.83 F1. This result when compared with the SciBERT-based neural model frustrates common expectations. Furthermore, this method is effective even w.r.t. the rate of semantification, since bioassays can be semantified in microseconds.

\section{Digital Library Bioassay Semantification Workflows}

We now describe the bioassay semantifier as an AI service application powering the structuring of scholarly knowledge in a real-world digital library (DL). The semantifier is importable within any DL that aim to establish knowledge-based information flows as the standard format for reporting and publishing research findings, aka contributions. The high-level workflow is a distributed, decentralized, and collaborative creation and development model comprising information templates, vocabularies, and ontologies (e.g., OBO foundry, Medline, MESH taxonomies, BAO in the Biomedical/Life Sciences domains). We discuss the service as implemented in TIB's Open Research Knowledge Graph (ORKG) platform (\url{https://www.orkg.org/})~\cite{auer_soren_2018}. The online semantification workflow will be a synergistic combination of automated and manual processes involving the extraction of new ontologized entity types from literature (e.g., target, assay type, experimental conditions in bioassays publications), open access data generation in accord with the FAIR principles thus easily reusable by anyone, and curation support tools for semantified data curation. Figures 2, 3, and 4 depict the workflow. It is pragmatically designed as a hybrid of automatic semantification linked to the BAO (\url{http://bioassayontology.org/}) and a simplified user interface to help scientists curate their data with minimum effort. This offers a highly accurate semantification model without placing unrealistic expectations on scientists to semantify their assays from scratch. In general, by thus drastically reducing the time required for scientists to annotate their contributions, we can realistically advocate for semantified contributions to become a standard part of the publication process. On such digitalized data, the ORKG additionally supports advanced data interlinking, integration, visualization, and search.

\begin{figure}[!tb]
  \centering
  \includegraphics[width=\textwidth]{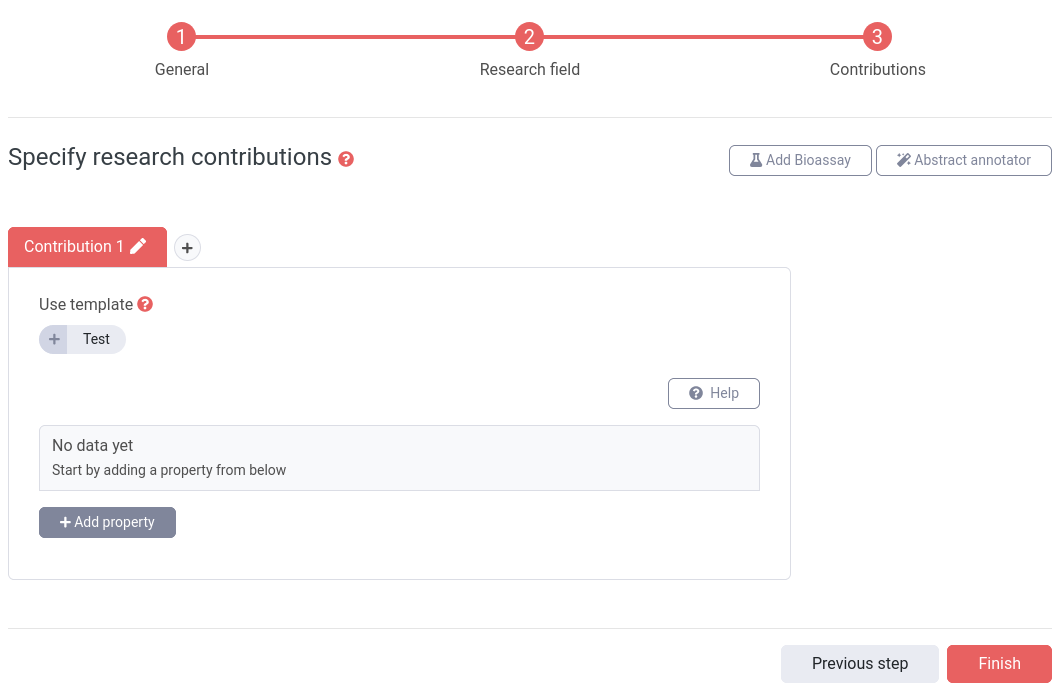}
  \caption{(1) General - add publication metadata; (2) Research field - select a research field from a taxonomy \url{https://gitlab.com/TIBHannover/orkg/orkg-backend/-/blob/master/scripts/ResearchFields.json}; and (3) Contributions - either structure an articles' contribution as \textit{method}, \textit{material} and \textit{results}, or add a bioassay text description by clicking `Add Bioassay.' \small{Note the `Add Bioassay' button is activated only for some research fields in the Life Sciences.}}
\end{figure}

\begin{figure}[!tb]
  \centering
  \includegraphics[width=0.8\textwidth]{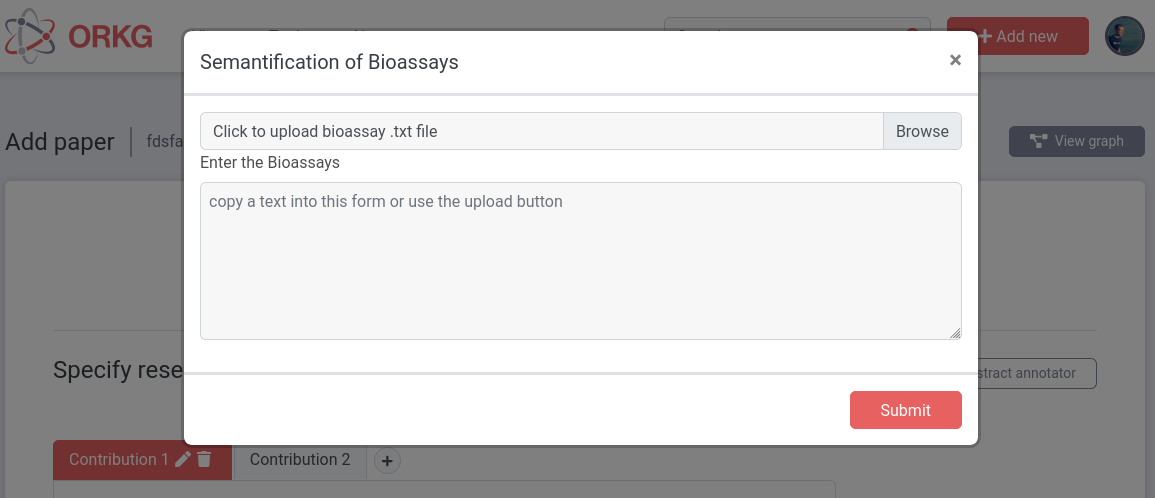}
  \caption{A popup pane to either upload or copy-paste a bioassay text description}
\end{figure}

\begin{figure}[!tb]
  \centering
  \includegraphics[width=\textwidth]{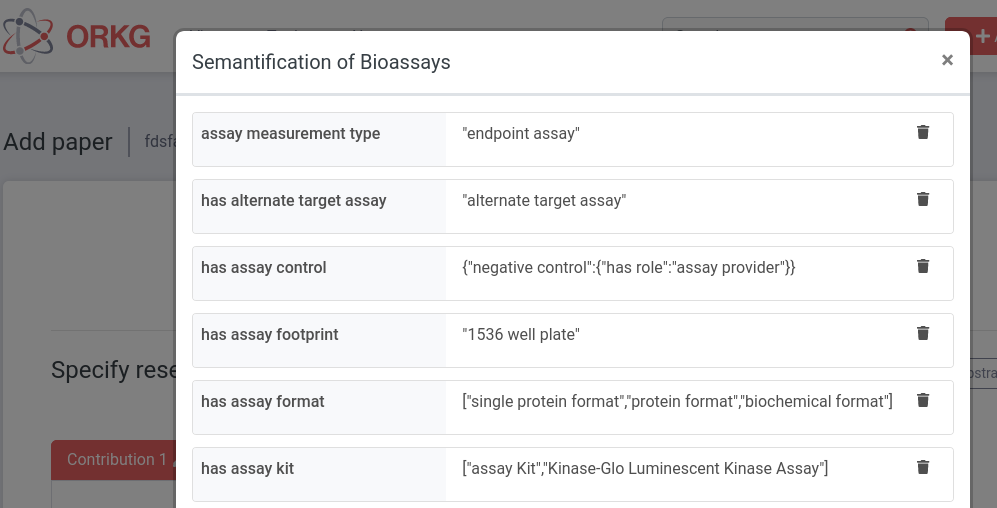}
  \caption{An automatically semantified bioassay based on submitted text with an interaction button to delete statements that the domain expert judges invalid}
\end{figure}

\section{Conclusion}

In this work, we have presented an end-to-end model to semantify bioassays descriptions in the context of knowledge-based digital libraries as the ORKG. As a result, we have implemented a highly accurate semantification machine learning method based on clustering. Our code is open source \url{https://gitlab.com/TIBHannover/orkg/orkg-bioassays-semantification}. Finally, we report an unconventional finding that resource-light clustering problem formulation can better support bioassay semantification than a state-of-the-art neural approach.

\clearpage

%
%
%
\bibliographystyle{splncs04}
\bibliography{mybib}
\clearpage

\appendix
\section{A Second Motivating Example for Bioassay Semantification}

\paragraph{\textbf{Assay ID 1061}} As another example sentence, we consider `The G-protein coupled formylpeptide receptor (FPR) was one of the originating members of the chemoattractant receptor superfamily. The present assay was undertaken to identify which of the 5 test compounds active in dose-response assays were FPRL1 antagonists.' While the sentence with various buried information units is not in a computable form, the logical statements `has participant' $\rightarrow$ `G protein coupled receptor' and `has role' $\rightarrow$ `target' about \textit{G-protein coupled formylpeptide receptor} are as the semantic equivalent of information buried in the text. Generally, in semantified bioassays, the predicates `has participant' and `has role' occur frequently as a related logical statement sequences. The label `has participant' often refers to a specific molecule participating in the bioassay, while the label `has role' refers to the role of the participating molecule in the experimental process. In the case of the discussed example, it means the bioassay has a `G protein coupled receptor' as a participant which is the \textit{target} molecule in the experiment. Note, similar to our previous example, this information is not explicitly found in the text and would rely on a human annotator observing context and their background knowledge of the experiment. In this paper our aim is to expedite the annotation process with the help of machine learning over representative annotated examples.

\section{Comparison of Reported Results with our Prior Work}
The dataset considered in this study has been created with a refined set of heuristics. Thus this dataset differs from the dataset in our earlier work. Note, our previous dataset had $|S|$ = 1756 unique statements (after filtering for non-informative ones), however, this dataset has 1906 unique statements. In the following lines, we describe the differences in our datasets and thus explain the new results reported in this paper.

\begin{enumerate}
\item The ignored classes differ as shown in Table~\ref{tab:classestoignore}. In addition in the old dataset we removed three specific labels: 'has function $->$ aggregated',
`has participant $->$ Calcium',
`has participant $->$ 7-amino-4-methylcoumarin'.

\item In the dataset that we use in the present study, we have combined all occurrences of the `has role' tag with its related tag pair. E.g. `has inducer \# has role $->$ Tetracycline \# inducer', `has participant \# has role $->$ Propionaldehyde
\# substrate'. 

\item previously we were including just the one with a BAO mapping ; here we have also other labels that are suitable for SciBert classification but are not in the BAO by checking the left hand manually and decided what to keep E.g. mapping: `http://www.w3.org/1999/02/22-rdf-syntax-ns\#type $->$ `10\% fetal bovine serum', label: `material entity culture serum' $->$ `10\% fetal bovine serum'.

\end{enumerate}

\begin{table}[]
    \centering
    \begin{tabular}{l|l|l|}\toprule
        Unique Statements LH & Old & New \\\midrule
        `has repetition point-number'& x & x\\
        `has concentration-point number'& x & x\\
        `has concentration value' & x & x \\
        `has endpoint'& x & x \\
        `has assay title'& x &x\\  
        `has quality' & x & \\
        `has mode of action' & x & \\
        `has concentration unit' & x & \\
        `has response unit' & x & \\
        `has inducer' & x & \\
        `antibody' & & x\\
        `has substrate' & & x\\
        `screening campaign name' && x\\
        `has transcription factor' & & x\\
        `material entity assay serum' & & x\\
        `material entity culture medium' && x\\
        `material entity culture serum'&& x\\
        `protein-protein'&& x\\
        `screening campaign name' && x\\
        `Annotated by'& & x\\
        `DMSO'&& x\\
        `NCBI taxonomy ID'&& x\\
        `PubChem TID'&& x\\
        `absorbance wavelength'&& x\\
        `cell modification temperature'&& x\\
        `cell modification time'&& x\\
        `construct DNA vector'&& x\\
        `construct artificial regulatory region copy number'&& x\\
        `construct gene ID' && x\\
        `construct organism'&& x\\
        `enzyme reaction time' && x\\
        `gene ID'&& x\\
         `gene mutation'&& x\\
         `has assay medium'&& x\\
         `absorbance wavelength'&& x\\
         `has emission wavelength value'&& x\\
         `has excitation wavelength value'&& x\\
         `has incubation time value'&& x\\
         `has signal direction'&& x\\
         `has summary assay'&& x\\
         `has temperature value'&& x\\
         `is alternate confirmatory assay of'&&x\\
         `is confirmatory assay of'&& x\\
         `is counter assay of'&& x\\
         `is identical assay of'&& x\\
         `is lead optimization assay of'&& x\\
         `is primary assay of'&& x\\
         `is selectivity assay of'&& x\\
         `positive control concentration'&& x\\
         `substrate incubation temperature'&& x\\
         `substrate incubation time'&& x\\
         `uniprot ID'&& x\\
        `material entity assay provider'& &x\\\bottomrule
        
    \end{tabular}
    \caption{Comparison among the old and the newly refined dataset. The Left-Hand parts of the labels are indicated.}
    \label{tab:classestoignore}
\end{table}
\begin{table}
\begin{small}
\parbox{.45\linewidth}{
\centering
\begin{small}
    \begin{tabular}{p{1.2cm}p{1.2cm}p{1.2cm}p{1cm}}\toprule
       $false$ labels & $P$ &  $R$ & $F1$  \\\midrule
       100 & 0.517 & 0.968 & 0.674 \\
       ... & ... & ... & ... \\
       160 & 0.549 & 0.931 & 0.688 \\
       \textbf{170} & \textbf{0.600} & \textbf{0.939} & \textbf{0.729} \\
       180 & 0.573 & 0.945 & 0.711 \\
       ... & ... & ... & ... \\
       300 & 0.471 & 0.674 & 0.551 \\\bottomrule
       \\
      \end{tabular}
\end{small}
    \caption{\footnotesize{Bioassay semantification results from five training optimization with different $false$ classification instances (full table in appendix)}}
      \label{tab:false}    
}
\hfill
\parbox{.45\linewidth}{
\centering
\begin{small}
    \begin{tabular}{p{1.8cm}p{1.2cm}p{1.2cm}p{0.9cm}}\toprule
       test set & $P$ &  $R$ & $F1$  \\\midrule
       1st fold & 0.600 & 0.939 & 0.729 \\
       2nd fold & 0.573 & 0.956 & 0.713 \\
       3rd fold & 0.589 & 0.936 & 0.719 \\
       \textbf{$Avg.$} & \textbf{0.588} & \textbf{0.944} & \textbf{0.720}
       \\\bottomrule
       \\
      \end{tabular}
\end{small}
    \caption{\footnotesize{Automatic bioassay semantification results from 3-fold cross validation with the optimal number of $false$ classification labels (170). 
    }}
      \label{tab:cv}    
}
\end{small}
\end{table}

\section{Classification Bioassay Results}

\begin{table}[!h]
    \centering
    \begin{tabular}{p{3.5cm}|p{2cm}|p{2cm}|p{2cm}}
    \toprule
          & P & R & F1 \\\midrule
         top 10, 170RF & \underline{\textbf{0.53}} & 0.94 & \underline{\textbf{0.67}}  \\
         top 20, 170RF & 0.50 & 0.89 & 0.64 \\
         top 30, 170RF & 0.45 & \underline{\textbf{0.95}} & 0.61 \\
         top 40, 170RF & 0.37 & 0.94 & 0.52 \\
         top 50, 170RF & 0.36 & 0.95 & 0.52 \\
         top 60, 170RF & 0.41 & 0.92 & 0.57 \\ 
         top 70, 170RF & 0.32 & 0.95 & 0.48 \\          
         full dataset, 170RF & 0.37 & 0.94 & 0.54 \\\end{tabular}

    \begin{tabular}{p{3.5cm}|p{2cm}|p{2cm}|p{2cm}}
    \toprule
         top 10, 180RF & \underline{\textbf{0.56}} & 0.93 & \underline{\textbf{0.70}}  \\
         top 20, 180RF & 0.50 & 0.93 & 0.64 \\
         top 30, 180RF &  0.49 & 0.94 & 0.65 \\
         top 40, 180RF & 0.35 &  \underline{\textbf{0.96}} & 0.51 \\
         top 50, 180RF & 0.39 & 0.94 & 0.55 \\
         top 60, 180RF & 0.40 & 0.95 & 0.56 \\ 
         top 70, 180RF & 0.37 &  0.95 & 0.53 \\          
         full dataset, 180RF & 0.35 & 0.94 & 0.51 \\\bottomrule
    \end{tabular}
    \caption{SCIBERT-based predictor results; 3-Fold CV on different subsets.  The first column contains the number of Random False (RF) or $false$ classification labels used in each bioassay for the analysed subsets (top 10, ..., full dataset) where each subset (e.g. top X) refers to the X most occurring Left-Hand (LH) part of a unique statement (e.g. top 10 contains the 10 most occurring LH of each unique statement). All the subsets where tested with 170 and 180 $false$ classification labels.  
    }
    \label{tab:results:SCIBERT:all}
\end{table}

\section{Elbow Optimization for K}

\begin{figure}[!h]
  \centering
  \includegraphics[width=0.7\textwidth]{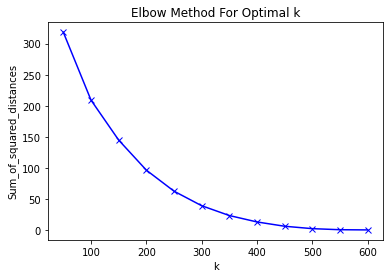}
  \caption{}
\end{figure}

\end{document}